# Visual analytics for team-based invasion sports with significant events and Markov reward process

*Kun Zhao; IBM Research - Tokyo; Tokyo, Japan.*
*Takayuki Osogami; IBM Research - Tokyo; Tokyo, Japan*
*Tetsuro Morimura; IBM Research - Tokyo; Tokyo, Japan*

## Abstract

*In team-based invasion sports such as soccer and basketball, analytics is important for teams to understand their performance and for audiences to understand matches better. The present work focuses on performing visual analytics to evaluate the value of any kind of event occurring in a sports match with a continuous parameter space. Here, the continuous parameter space involves the time, location, score, and other parameters. Because the spatiotemporal data used in such analytics is a low-level representation and has a very large size, however, traditional analytics may need to discretize the continuous parameter space (e.g., subdivide the playing area) or use a local feature to limit the analysis to specific events (e.g., only shots). These approaches make evaluation impossible for any kind of event with a continuous parameter space. To solve this problem, we consider a whole match as a Markov chain of significant events, so that event values can be estimated with a continuous parameter space by solving the Markov chain with a machine learning model. The significant events are first extracted by considering the time-varying distribution of players to represent the whole match. Then, the extracted events are redefined as different states with the continuous parameter space and built as a Markov chain so that a Markov reward process can be applied. Finally, the Markov reward process is solved by a customized fitted-value iteration algorithm so that the event values with the continuous parameter space can be predicted by a regression model. As a result, the event values can be visually inspected over the whole playing field under arbitrary given conditions. Experimental results with real soccer data show the effectiveness of the proposed system.*

## Introduction

Team-based invasion sports include soccer, basketball, and hockey [5]. A common feature of these sports is that two teams compete to "invade" the opponent's end of the field and score by shooting the ball into the opponent's goal. Efficient visual analytics can provide audiences and team managers with better understanding of match. For example, there are generally few goals during a 90-minute soccer match, but if we can show the importance of various events in contributing to a goal, then audiences can understand the match better, and team managers can make better decisions for the team.

Spatiotemporal data, meaning the trajectories and event logs of a match, is generally used in such analytics. The trajectory data might be captured automatically by camera sensors, including the trajectories of each player, the ball, and the referee. Each second of trajectory data contains 15-30 frames. On the other hand, event logs contain events occurring throughout the match, such as passes, fouls, throw-ins, and so on. Both trajectories and event logs are low-level representations of the match and have a very large data size [5][8].

In this work, we focus on performing board evaluation, like that shown in Figure 6, of events with a continuous parameter space. Here, board evaluation refers to a visual analytics approach that can map the event value for any event type to the whole playing field with a continuous parameter space involving an arbitrary time, an arbitrary score, and so on. In real sports events can occur anywhere and anytime, and board evaluation of events with a continuous parameter space enables evaluation of event values under any given conditions. As a result, the main challenge, or requirement, in achieving such board evaluation can be summarized as the following:

- **R1**: Deal with the complexity of the continuous parameter space so that event values can be estimated under arbitrary conditions.

The event type is actually in a discrete space but can also be solved as a continuous-parameter problem. We hereinafter also regard the event type as one parameter in the continuous parameter space. Because spatiotemporal data is a low-level representation and has a very large size, however, traditional analytics systems may need to discretize the continuous parameter space by subdividing the playing area [7][11][15][17][19] or limit the analysis to certain specific events or local performance by using some local-area-based features [3][18][19][21][28]. To the best of our knowledge, there is still no visual analytics system for board evaluation of events with a continuous parameter space.

To deal with the complexity of the continuous parameter space, we consider a match as a Markov chain of major events (i.e., intensity-based significant events). Then, event values with the continuous parameter space can be estimated by solving the Markov chain with a machine learning model. Here, a Markov chain [25] refers to a chain of states in which the probability of each state depends only on the previous state. This property is called the Markov property. During a soccer match, many events occur (e.g., a foul, ball out of bounds, etc.). Some events are less important, while others might be more important in influencing the match. Direct usage of all events as states might lead the chain to violate the Markov property. For example, suppose that we have an event sequence of corner kick – pass – pass – shoot. Here, a pass event is followed by a shoot event, but it is hard to conclude that the Markov property applies between these events. Actually, it is natural to consider the shoot event as a contribution of the corner kick event. As a result, to maintain the Markov property, we must extract events that can abstract the play of the whole match. This yields another requirement:

- **R2**: Extract major events that can abstract the whole match to build a Markov chain.

With these two requirements in mind, our analytics approach proceeds as follows.

First, we extract major events that can abstract the whole match to build the Markov chain and thus meet the requirement of R2. We hereinafter call such events "significant events". Considering the phases of a soccer match, we extract two types of events: in-play-type significant events for in-play phases, and stoppage-type significant events for stoppage phases. For the in-play type, we extract significant events according to the playing intensity. We propose a new feature extractor using a multivariate distribution of the players to detect the playing intensity and extract the significant events. On the other hand, we also extract events after a relatively long stoppage as stoppage-type significant events. These events cannot be extracted using playing intensity but can directly contribute to scoring (e.g., a penalty kick) and can be considered as major events to abstract the whole match.

Second, the extracted significant events are redefined as different states with a continuous parameter space. Using the continuous parameter space for the state definition ensures that value estimation is performed in consideration of the parameters. A Markov chain of the redefined states is then built so that a Markov reward process [9] can be applied. For each match, we build a Markov chain to represent the match.

Third, a normal Markov reward process can only solve a problem with a limited number of states. We thus introduce a customized fitted-value iteration algorithm with a regression model to solve the Markov reward process. Because the continuous space is directly used as an input to the regression model, estimation of event values under arbitrary conditions becomes possible. We designed the second step and this step to meet the requirement of R1, because the continuous parameter space is input and used to estimate event values.

With the above three steps, we can estimate and visualize the value of any event type with the continuous parameter space. Event values can be visually inspected over the whole playing field to understand how much each event can potentially contribute to scoring, without limitation to a specific area subdivision, event type, or situation.

To verify the effectiveness of the proposed system, we applied it to spatiotemporal soccer data of the J1 League (the top division of the Japan Professional Football League). We first used the proposed feature extractor to extract intense periods of matches and then compared them to manually extracted highlights to verify whether the intense periods were extracted properly. Then, we calculated event values with the proposed system and mapped them to the soccer field so that event values under any different conditions could be visually inspected. Feedback from domain experts also verified the effectiveness of our proposed method.

To sum up, the contributions of this work include the following:
- We propose to use a Markov reward process to estimate event values so that they can be calculated with a continuous parameter space (R1).
- We introduce a new feature extractor to extract significant events according to the playing intensity and use them to build a Markov chain (R2).
- We visually inspect event values over the whole playing field to provide a deeper understanding of the value estimation results. Application to a real soccer data set and feedback from domain experts verify the effectiveness of the proposed system.

## Related Works

With the development of tracking techniques, trajectories and event logs can be found for many team-based invasion sports. This kind of spatiotemporal data is a low-level representation of a sports match and has a large size [5], making sports analytics using such data a challenging problem. In this paper, we focus on visual analytics to estimate event values with a continuous parameter space.

For the requirement to deal with the complexity of the continuous parameter space (R1), one popular approach is to discretize the spatial complexity. For example, a continuous space can be subdivided into regions, and then region-based features can be extracted for analytics. Camerino et al. [14] proposed extracting patterns from a soccer match by subdividing the playing area into 15 regions based on the lateral position (right, center, and left) and zone (ultra-defensive, defensive, central, offensive, and ultra-offensive). They also combined nine patterns according to the ball location to increase the representational ability of the extracted patterns. Bialkowski et al. [7] also proposed a rectangle-based subdivision of the playing area for soccer, using rectangles of the same side length. A similar subdivision of the playing area was used by Cervone et al. [11] for basketball. Shortridge et al. also subdivided the playing area for basketball analytics [15]. Such subdivision of the playing area is very common for team sports and can be found in many other related works [16][17][19][24]. When the playing area is subdivided, the low-level features (i.e., trajectories, and event logs) can be abstracted into region-based features, and analytics can be performed much more easily. No matter how well the playing area is subdivided, however, local differences between slightly different locations might be neglected. To meet the requirement of R2, instead of subdividing the playing area, we keep the original continuous space so that spatial features can be reflected in the evaluation results. We also take the same approach of keeping the original continuous parameter space for other parameters.

To keep the original continuous parameter space, we need a method to deal with its complexity so that event values can be estimated with arbitrary conditions. In the development of artificial intelligence, machine learning is considered a good solution for estimating values in a continuous space. Matthew et al. [22] used deep reinforcement learning for estimation in a continuous action space. Warwick et al. [23] introduced a reinforcement learning algorithm for Markov decision processes with parameterized actions. A similar approach can also be found in [26]. All those works mainly focused, however, on the learning algorithm to approximate the value function or Q-function in a continuous space, without much consideration for the features of sports. Moreover, those works sought to estimate the action space, but our work seeks to evaluate the value space (i.e., the space of event values) without considering the action space.

In this paper, we also introduce machine learning to solve the continuous parameter space problem. Instead of using a normal value function or Q-function (as in the above works [22][23][26]), we propose to use a Markov reward process to estimate event values without considering the action space. We build a Markov chain for each match and seek to extract match-based features to define states so that the Markov property can be maintained for the built Markov chain (R2).

There are also many works related to sports feature extraction. As a simplest approach to extract features from a match, one work proposed to roughly sample small chunks of equal length as segmentation events to represent the match [2]. This rough method may neglect some major events that are highly related to the score, and it is thus unsuitable for maintaining the Markov property. Fernandez et al. [18] proposed to use each player's temporal space occupation as a feature to analyze the player's performance. To evaluate attacking performance in soccer, Link et al. [19] defined a notion of "dangerousity," which is present for every moment in which a player possesses the ball and can therefore complete an action with it. Gudmundsson et al. [21] extracted a dominant region based on the movement of each player. Lucey et al. [20] proposed to use a 10-second window of play before a player's shot to evaluate the shot performance. Analytics studies focused on passing in soccer can also be found in [12][28]. The common point of those works is that features are extracted according to the local performance of some players, or that the analytics are limited to some specific events (e.g., a shot, pass, etc.). Because we regard the whole match as a Markov chain, our system extracts features by considering the whole match to maintain the Markov property. We extract significant events according to the intensity of the players' activity, because we think that such events contribute more strongly to scoring goals. In other words, such events better enable maintenance of the Markov property.

## Team sports board evaluation system

In this section, we briefly describe the proposed team sports board evaluation system. With the two previously mentioned challenges (i.e., global feature extraction and value estimation with a continuous space) in mind, this system can be separated into two primary stages: significant event extraction and value estimation.

### *Significant event extraction*

Trajectories and event logs constitute a low-level representation of a sports match. In this subsection, we explain how to use significant events that may contribute to scores, like those shown in Figure 1, as high-level global features to abstract and represent a whole match. We consider the playing intensity as a criterion to detect significant events.

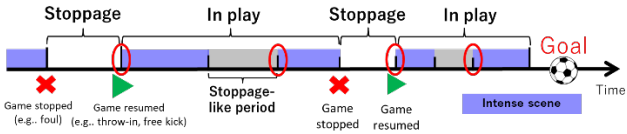

*Figure 1. Significant events used in the proposed method, which are defined as those occurring after a reset period. The red circles indicate significant events.*

Because the situations of in-play phases and stoppages differ, significant events are extracted from them with different methods.

During an in-play phase, we think that periods with high playing intensity are more likely to influence the match, especially to influence scoring, and can thus be used to abstract the match. For example, when one team approaches the goal area, both the offensive and defensive teams move rapidly to attack and defend. We call such a period an "intense period," as shown in Figure 2. These intense periods are likely related to scoring and can be used to abstract the whole match. We extract the start of an intense period as an **in-play-type significant event**. In fact, a match does not always include intense periods, because much of the match time is used for passing the ball within a team to find chances to set up an attack. We call such periods "stoppage-like periods," as shown in Figure 2. These stoppage-like periods are not used to abstract the whole match. During a stoppage-like period, players are supposed to reset their positions to proper locations to prepare to attack or defend.

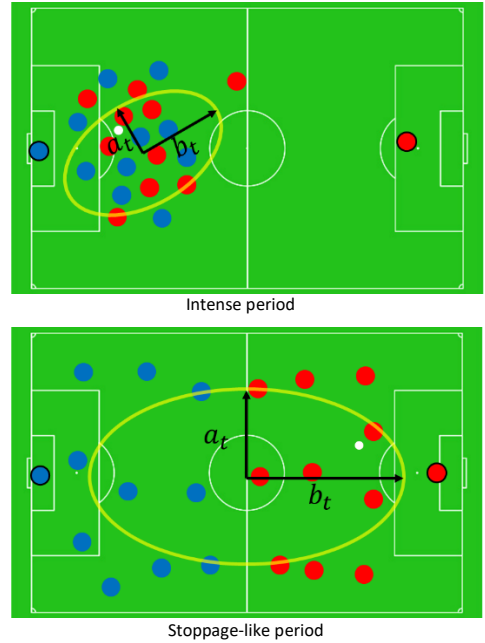

*Figure 2. Examples of an intense period (top) and a stoppage-like period during an in-play phase (bottom). The blue and red circles represent the respective players of the two teams. The white circle represents the ball. The top image shows a period in which the red team is getting ready to shoot. The bottom image shows a period in which the red team is passing the ball on their own side, and players are supposed to be in their proper positions. The yellow ellipses represent the covariances of the player distributions, with $(a_t, b_t)$ indicating the eigenvalues of the covariance matrix.*

When a match is in a stoppage phase, we define a significant event as an event after a stoppage (e.g., a free kick after a foul). During such a stoppage, both defenders and attackers are supposed to come back to their proper positions if the stoppage period is sufficiently long. We hereinafter refer to this kind of stoppage as a "reset period," because players are supposed to reset to their proper positions to set up the attack or defend. Such events might not be captured by analyzing the playing intensity (e.g., a penalty kick or direct free kick can directly score a goal but is not accompanied by an intense period), but they are important for building the Markov chain. We refer hereinafter to these events after a reset period as **stoppage-type significant events**.

Stoppage-type significant events can easily be extracted from event logs. In-play-type significant events, however, should be extracted from the intensity of in-play phases. In fact, this is very challenging, because the intensity of a sports match is very hard to define. In one related work [2], an in-play phase was segmented into small chunks of equal length, and these events were used to represent the in-play phase. This is a very rough method and may neglect some major periods that are highly related to scoring.

### Detection of intense periods

In the present work, we use the trajectories of the players to extract intense periods during an in-play phase. We assume the players in the playing area have a multivariate distribution (e.g., a multivariate Gaussian distribution) and use the covariance of the players' locations to detect the intensity. Generally, if we have a period in which the offensive team is passing the ball within their own side (i.e., a stoppage-like period), so that both the offensive and defensive teams are in their proper positions to prepare for the next attack, the players are supposed to be dispersed widely, and a large covariance can be detected for the distribution. In contrast, if we have a period in which the offensive team is approaching the goal area and preparing to shoot (i.e., an intense period), players of both the offensive and defensive teams are close to the ball to attack and defend, respectively, so that a small covariance can be detected for the distribution. As a result, the change from a stoppage-like period to an intense period can result in an obvious change in the distribution covariance. It is also not hard to see that, if the offensive team sets up a play to approach the goal area, players of both teams may have very active movement, and the distribution covariance should also change a lot in this case. In this work, we thus extract features from changes in the distribution covariance to detect intense periods. The detailed process is listed as follows.

I. For an in-play phase, assume that players are distributed according to a multivariate probability distribution (e.g., Gaussian, Laplace etc.). Calculate the covariance matrix $\Sigma_t$ at time $t$:

$$\Sigma_t = \begin{pmatrix} \sigma_{xx} & \sigma_{xy} \\ \sigma_{xy} & \sigma_{yy} \end{pmatrix}, \quad (1)$$

where $x$, $y$ denote the player coordinates of the two teams, and $\sigma$ denotes the variance.

II. Calculate the eigenvalues ($a_t$, $b_t$) of the above covariance matrix (shown by the vectors in Figure 2), and calculate the area $S$ of the corresponding ellipse as

$$S(t) = \pi a_t b_t. \quad (2)$$

The area $S$ is used as a feature to represent the intensity of the match. Then, calculate the absolute value of the difference, $f(t)$, as follows.

$$f(t) = |S(t) - S(t-1)| \quad (3)$$

III. Define a function $g(t)$ as

$$g(t) = \begin{cases} 1, & -N \leq t \leq 0 \\ 0, & else \end{cases}. \quad (4)$$

Then calculate the convolution as follows, where $n$ is the integration length:

$$(f * g)(t) = \frac{1}{n} \sum_n f(\tau) g(t - \tau). \quad (5)$$

IV. Find local peaks over a range of $N$ seconds with the following condition:

$$T = \{t | (f * g)(t) > (f * g)(i) \}, i = [t-N, t+N], i \neq t. \quad (6)$$

Such peaks for $t \in T$ in the range of $[t-N, t+N]$ are extracted as intense periods.

Here, $N$ is a parameter to control the length for extracting intense periods. This parameter can be optimized by a validation process.

The above Step II calculates the change in the distribution covariance. Direct use of this instantaneous value, however, is not suitable for extracting the intense periods, because it may contain noise due to sensor errors in the tracking system. This is why we calculate the convolution of this instantaneous value in step III to obtain the amount of change, so that intense periods can be expressed. After this, we also need to determine the exact start and end times of the intense periods. This is performed in step IV, which finds local peaks over a given length of 2$N$ and extracts periods covering the local peaks as intense periods. Here, if we used global peaks to extract the intense periods, we would only extract a few periods, such as those before a shot or directly related to a goal, because such periods have the highest convolution values. For a sport like soccer, there may be few scores during a whole match, and global peaks might also be very few. Instead, the usage of local peaks is efficient for extracting various kinds of intense periods, including those during normal play.

As a result, the covariance-based area provides a criterion to detect the intensity, and the start of an intense period is extracted as an in-play-type significant event.

*Value estimation*

In this subsection, we first define the state based on the event type, location, time, and other conditions. Then, a Markov chain of states is built according to state transition probabilities. A Markov reward process is then applied to the Markov chain by considering a goal as a reward. We also introduce a customized fitted-value iteration algorithm with a regression model to solve the Markov reward process. Because the continuous space is directly used as the input of the regression model, it becomes possible to estimate event values with arbitrary conditions. As a result, the proposed system can estimate the value of an event anytime, anywhere, under any other condition.

**State definition**

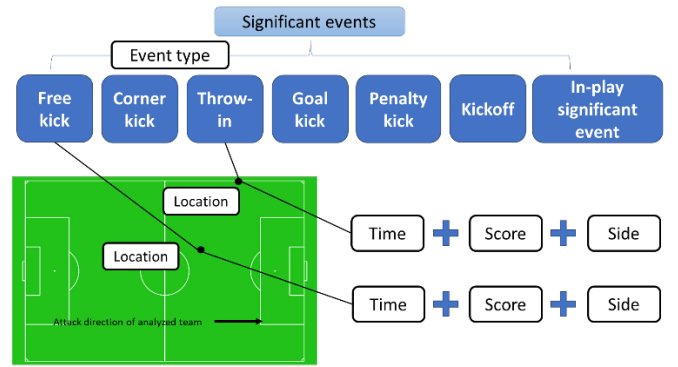

State = Time + Event type + (x, y) + Self/Opponent + Self score + Opponent score

Figure 3. *Definition* of the state in terms of the event type, position, time, score, and team side.

Different significant events can be roughly used to estimate event values. This approach does not, however, consider a detailed parameter space including parameters such as the time, place, score, and other conditions. Because we focus on visually inspecting event values under different conditions, we need to define the state in terms of not only the event type but also other parameters. In this work, we use the event type, location, time, team (home or away), and current score, which can all influence the scoring of a goal, as features to define the state, as shown in Figure 3.

Because the parameters such as location and time are in a continuous space, the state definition also results in different states. The continuous parameter space is thus kept in each state, which is used in value estimation.

**Markov reward process**

With the states defined, we assume that different states can form a state transition diagram like that in Figure 4. For each half of a match, a time-ordered Markov chain of states can be built according to the state transitions. Figure 5 shows an example of a Markov chain of events. In this paper, we define the event value $v$ as the **goal difference until the end of the match**, with consideration of a discount factor $\gamma$:

$$v(X) = E[\sum_t \gamma r_t]. \qquad (7)$$

Here, $r_t$ refers to the reward, which is defined as +1 when the analyzed team scores a goal and as -1 when the opposing team scores a goal. Therefore, we can define the following Markov reward process:

$$\boldsymbol{v} = \boldsymbol{r} + \gamma \boldsymbol{P}\boldsymbol{v}. \qquad (8)$$

Here, $\boldsymbol{r}$ refers to the reward vector, and Figure 5 also shows examples of $\boldsymbol{v}$ and $\boldsymbol{r}$. $\boldsymbol{P}$ refers to the transition probability matrix between different events and can be derived for a given dataset. The discount factor $\gamma$ ranges between 0 and 1. Equation (8) can also be transformed to (9) so that the value of $\boldsymbol{v}$ can be calculated.

$$\boldsymbol{v} = (\boldsymbol{I} - \gamma\boldsymbol{P})^{-1}\boldsymbol{r} \qquad (9)$$

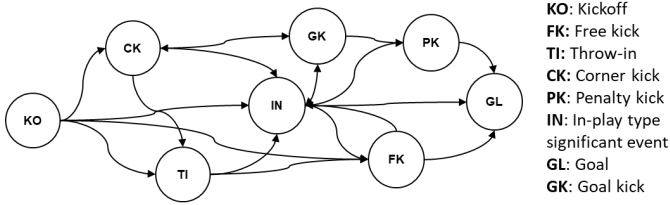

**KO**: Kickoff
**FK**: Free kick
**TI**: Throw-in
**CK**: Corner kick
**PK**: Penalty kick
**IN**: In-play type significant event
**GL**: Goal
**GK**: Goal kick

Figure 4. Example of a state transition diagram. We use the event types to represent different states.

**Fitted-value iteration**

Because the state definition keeps the continuous parameter space, the number of states is infinite. A normal Markov reward process like that of equation (9) cannot solve a problem with infinite states. Simply speaking, if there is an infinite number of states, then the value vector $\boldsymbol{v}$ also has infinite length, making equation (9) impossible to solve. In fact, the complexity of the continuous parameter space is a reason why some related works had to discretize the playing area (e.g., [7][11][14]). Because we focus on board evaluation, discretization is not suitable. To solve this problem, we thus propose to use a fitted-value iteration algorithm to estimate event values with a continuous parameter space.

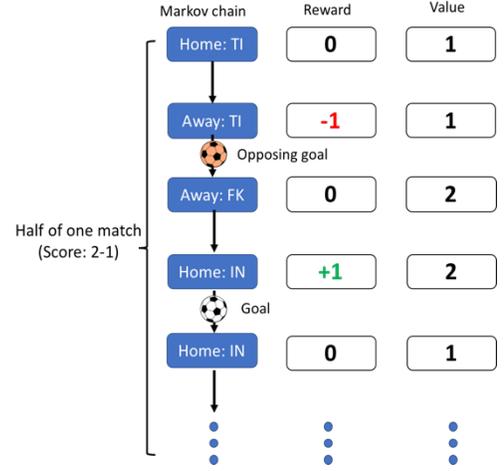

Figure 5. Example of a Markov chain of states, with rewards and values, for a match ending with a score of 2-1 in favor of the analyzed team. Each node refers to a state. We use the event types to represent different states. TI denotes a throw-in, FK denotes a free kick, and IN denotes an in-play-type significant event. The values shown in this example use a discount factor of 1.

By using a value iteration process [10], for a given time $t$ the Markov reward process in (8) can be rewritten as a value iteration formula:

$$v_t \leftarrow r + \gamma v_{t+1}, \qquad (9)$$

where $r$ is the state's reward. The example in Figure 5 shows a case with a discount factor of 1. We then import a regression model $\emptyset$ and assume the value $v$ can be represented as

$$v = \emptyset(X). \qquad (10)$$

Here, $X$ represents the state features shown in Figure 3, defined as the event type $e$, location $l$, time $t$, score $s$, and team $h$. Then, by substituting (9) into (10), we have

$$\emptyset(X_t) \leftarrow r + \gamma\emptyset(X_{t+1}), \qquad (11)$$

where $X_t$ denotes the state features at time $t$. As a result, given match data, we can perform this process for each consecutive state to train the regression model $\emptyset$.

Letting $v(X)$ denote the value of a significant event with feature $X$, we can summarize the fitted-value iteration algorithm as follows. The immediate reward $r(e_t)$ is given by

$$r(s) = \begin{cases} 1 & \text{if the event results in a score} \\ -1 & \text{if the event results in an opposing score} \\ 0 & \text{otherwise} \end{cases}$$

Using Algorithm 1, we can train the regression model to estimate the expected value for an arbitrary given set of state features.

---

Algorithm 1: Fitted-value iteration

```
1: Input:  Data $\Gamma \equiv \{X \equiv (e, l, i, s, h)\}$; Maximum iterations
   $N$; Regression model $\emptyset$
2: Initialize values: $v(X) \leftarrow r(e)$
3: for $n \leftarrow 0$ to $N$ do
4:     Prepare dataset $D$ for regression: $D \leftarrow \{(X, v(X)\}$
5:     Fit $\emptyset$ with $D$
6:     for $X_t \in \Gamma$ do
7:         if $X_t$ is the last state feature:
8:             $v(X_t) \leftarrow 0$
9:         else
10:            $v(X_t) \leftarrow r(e_t) + \gamma\emptyset(X_{t+1})$
11:        end if
12:    end for
13: end for
```

## Experimental results

In this section, we apply our system to a real soccer dataset for the J1 League in the 2015 season. This data includes both sensor data, containing the trajectories of all players and the ball, and event logs, which are created manually to record events such as kickoffs, passes, and so forth. In total, there are 611 matches in the dataset.

We first used the proposed feature extractor to extract the intense periods of a match and compared them to manually extracted highlights to verify whether the intense periods were extracted properly. Then, we used the fitted-value iteration algorithm to calculate event values and map them to the soccer field so that the event values under any different conditions could be visually inspected.

### Extraction of intense periods

Sports highlights are defined as the abstractions of sports to summarize an entire match [13]. In our proposed system, intense periods are also extracted to abstract the whole match and also can be used as the sports highlights. Given a sports match data, highlights are normally selected and created by a human [2]. An automatic method of extracting highlights would alleviate this burden from a human, but automatic highlight extraction is a challenging and important problem. As a result, we can compare our extracted intense periods with manually extracted highlights to verify whether intense periods are extracted correctly.

As an experiment, we tested the match between Shimizu S-Pulse and Gamba Osaka on April 12th, 2015. The extracted intense periods were compared with manually selected highlights to calculate the accuracy of extraction. The length of an intense period was set to 20 seconds. The manually selected highlights, however, had different lengths for different periods. We counted an intense period as correct if it appeared in the manually extracted highlights. To provide a comparison baseline, we took the convolution of the average speeds of all players and extracted local peaks as highlights, because player speed is also supposed to be an important factor related to highlights.

As a result, our proposed method detected 138 highlights. We ranked the highlights by convolution value and found that, among the top 40 highlights, 33 could also be found in the manually created highlights, giving 82.5% accuracy. In contrast, the baseline method detected 202 highlights, and the top 40 had only 37.5% accuracy. These results, shown in Figure 6, confirmed that the proposed method could provide much higher accuracy.

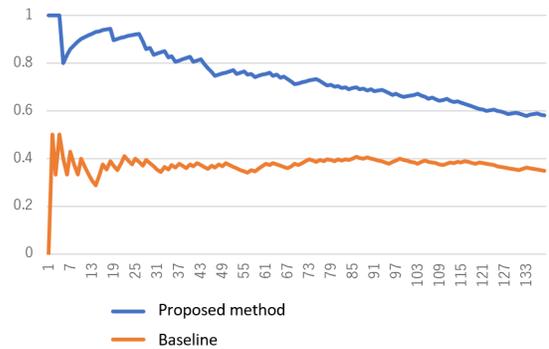

Figure 6. Accuracy of the extracted highlights (intense periods) with our proposed method and the baseline (highlights extracted by using players' speed). The detected highlights were ranked by the convolution value calculated with equation (5). The horizontal axis indicates how many highlights in convolution order were used to calculate the accuracy.

We also found that the whole set of 138 automatically detected highlights covered 85.1% (80/94) of the highlights selected by a human. The detected highlights included not only the periods before goals but also scrambles, fouls, and other kinds of intense periods. This high accuracy provided good assurance for the analytics described in the following subsection.

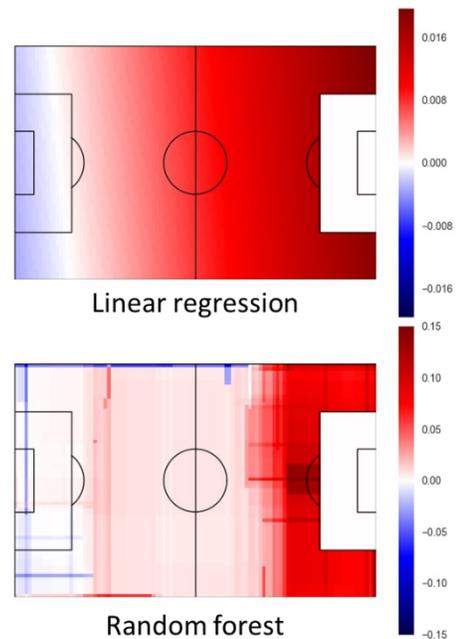

Figure 7. Value of a direct free kick at 8:20 with a score of 0-0. The top image is the calculation result with the linear regression model, while the bottom image is the result with the random forest model. The results are oriented so that the direction of offensive play is from left to right.

## Value calculation with fitted-value iteration

From the J1 League soccer data, 202,168 significant events were extracted. Each match was used as one episode. While 70% of this data was used for training, the remaining 30% was used for validation of the accuracy. For the regression model, we tested both a linear model (linear regression) and a nonlinear model (random forest). The calculated values were visually inspected by generating a heat map of the corresponding locations on the soccer field. Figure 7 shows the value of a direct free kick at a time of 8:20 with a score of 0-0 for both models. These and all subsequent visualization results are oriented so that the direction of offensive play is from left to right. Here, the discount factor was set to 1.0 to give each goal a longer influence, because there are generally few goals in a soccer match. The color maps have a range of $[-M, M]$, where $M$ was the maximum absolute event value.

We also measured the root mean square error (RMSE) of each model with the validation data, as defined by the following equation:

$$RMSE = \sqrt{\frac{1}{n}\sum_{i=1}^{n}(\emptyset(X_i) - V_i)^2}. \qquad (12)$$

Here, $n$ denotes the number of samples. The linear regression model generated an error of 0.8586, while the random forest model generated an error of 0.8413. Here, the error was calculated as the RMSE between the predicted event value and the real value (integrated number of goals). We can see that the random forest provided better performance than that of the linear regression. Besides the numerical results, it is also obvious that the nonlinear random forest model can provide more detailed results for the value distribution. As a result, the following analytics were performed with the random forest model.

Figure 8 shows the results for a direct free kick at 8:20 but with various scores. From the results we can observe that the right side (the offensive side) has relatively high values, matching the fact that locations near the goal area provide a better chance for a direct free kick to induce a goal. Moreover, the area with the highest values is near the penalty arc, which is the nearest area to the goal for a free kick. Another interesting finding is that the far end of the left side can have negative values. This indicates that, if a direct free kick is in the area near a team's own goal, the team is more likely to give up a goal. The reason might be that if the direct free kick is too far from the goal, the opposing team can steal the ball and counterattack.

We also evaluated the values of different event types, as shown in Figure 9. Here, all the events were assumed to occur at 8:20 with a score of 0-0. From these results we can see that an indirect free kick has a similarly high value distribution for the offensive team. The detailed pattern is different, however, in that an indirect free kick tends to have its highest values near the center circle on the offensive side. This is not difficult to understand, as the indirect free kick is not for direct shooting but for setting up an attack, and the above-mentioned area tends to be the best area for such setup. Another difference is that the direct free kick has negative values for areas near the offensive team's own goal, while the indirect free kick has positive values for the same areas.

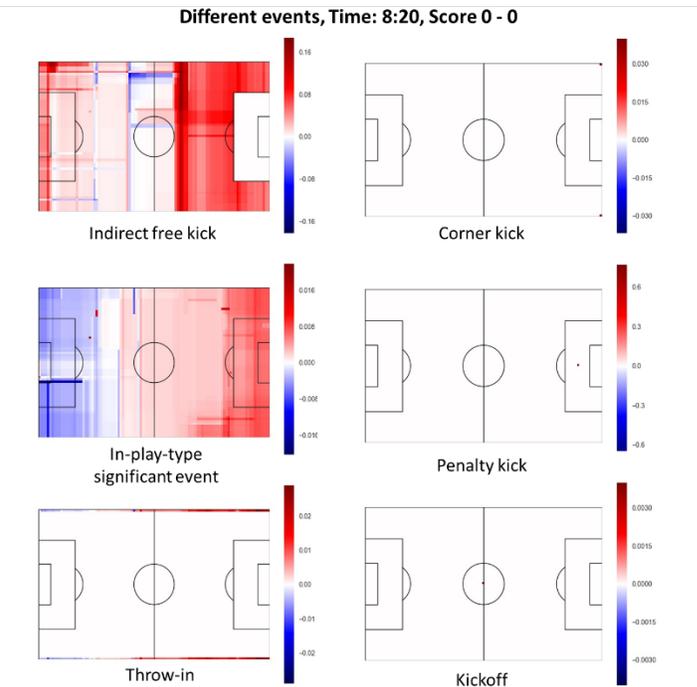

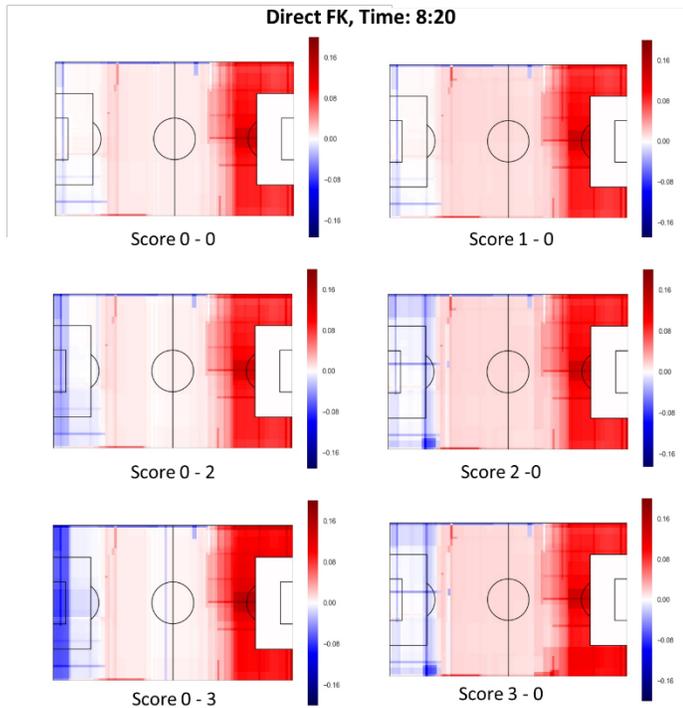

*Figure 8. Board evaluation of a direct free kick at 8:20 with different scores (own score – opposing score).*

*Figure 9. Board evaluations of different kinds of events occurring at 8:20 with a score of 0-0. Here, the corner kick has a value around 0.03 on both sides, the penalty kick has a value over 0.7, and the kickoff has a value very near 0.*

Figure 9 also shows a board evaluation of in-play-type significant events. The event location was extracted as the ball location at the start of the intense period. From this result we can see that areas near the offensive side have positive values, while areas far from the offensive side have negative values. Because the event location is the ball location at the start of the intense period, it appears that the positive-valued area corresponds to the beginning of an attack, when both teams' play becomes intense. Such periods can contribute to scoring. On the other hand, the negative-valued area might represent that the opposing team may steal the ball and counterattack, which might cause the analyzed team to give up a goal.

As for the other events shown in Figure 9, the evaluation result for a throw-in indicates that locations near the offensive side may offer a better chance to score a goal than locations farther away. Corner kicks from both sides show a positive value around 0.03. A penalty kick shows a very high value over 0.7, which also indicates the accuracy of our analysis results, because a penalty kick is generally a very good chance to score. As for a kickoff, the value is near zero, indicating its very limited influence on scoring.

Time is also an important factor for teams to score. Figure 10 shows the board evaluation results for a direct free kick at different times. From the figure we can see that as time passes, the color becomes lighter and lighter, indicating that the event values become lower and lower. This is because, as the remining time becomes less and less, the chance of scoring a goal also becomes smaller and smaller. We can also observe that the highest-value area near the penalty arc at 8:20 becomes more similar in value to areas near it over time.

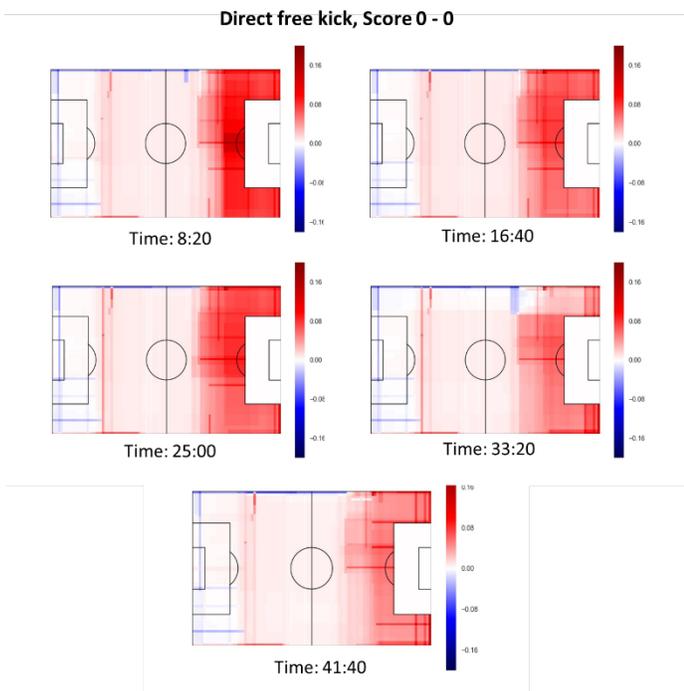

Figure 10. Board evaluations of direct free kicks at different times.

## Discussion

From the above results we conclude that our system can provide efficient visual analytics by considering a continuous parameter space including the location, time, and other factors. We also showed these results to domain experts. They gave positive feedback and believed that such results would be helpful for audiences to understand soccer matches better and for team managers to analyze and improve team performance. The proposed intense-period extraction technique, which can be used to extract the highlights of a match, was also considered a good method for automatic highlight detection. Highlight extraction is generally performed by a human, and an automatic method of extracting highlights would alleviate this burden from a human. Note that, because the analytics were performed for all matches through the 2015-2016 season of the J1 League, the results show trends over the whole league rather than a specific team. As a future work, we will focus our research on a specific team and perform more detailed analytics. We also plan to add other important features to define states, enabling a more complicated value estimation based on different conditions.

On the other hand, the domain experts also pointed out that some areas showed strange event values. For example, in Figure 10, the results at 33:20 show that a direct free kick in the area near the top center has a negative value, represented by an obvious blue area. According to the domain experts, this area should also be a good position for a team to set up an attack, so the event value should be positive. In fact, this might have been caused by overfitting of the training data, which can be solved with a larger training dataset.

Moreover, in this work we have used a linear regression and random forest as regression models. Application of a deep neural network in our system is also possible. In fact, usage of a deep neural network, like a deep-Q network (DQN) [27], can be found in many other works to calculate results for a continuous parameter space. Generally, deep neural networks can increase the descriptive ability of a model but also have disadvantages like a long training time, complex structure, and difficult parameter tuning. We will consider combining a deep neural network with our analytics approach in the future.

## Conclusion

In this paper, we have proposed a visual analytics system for soccer by using intensity-based significant events and a Markov reward process. The significant events are first extracted by considering the playing intensity during a match. Then, the extracted events are redefined as different states with a continuous parameter space and built as a Markov chain so that the Markov reward process can be applied. Finally, the Markov reward process is solved by a customized fitted-value iteration algorithm so that the values of events can be estimated in the continuous parameter space. As a result, the event values can be visually inspected over the whole playing field under any given conditions. Experimental results with real soccer data showed the effectiveness of the proposed system.

## Acknowledgments

We thank Data Stadium Inc. for providing the J1 League soccer dataset. This work was supported by JST CREST Grant Number JPMJCR1304 in Japan.